\pdfoutput=1
\documentclass{article}

\usepackage[nonatbib]{neurips_2019}

\usepackage[utf8]{inputenc} 
\usepackage[T1]{fontenc}    
\usepackage{hyperref}       
\usepackage{url}            
\usepackage{booktabs}       
\usepackage{amsfonts}       
\usepackage{nicefrac}       
\usepackage{microtype}      
\usepackage{amsmath,amssymb}
\usepackage{mathtools}
\usepackage{bbm}
\usepackage{verbatim}

\newtheorem{theorem}{Theorem}

\newtheorem{lemma}{Lemma}

\def\H{\mathcal{H}}
\def\O{\mathcal{O}}
\def\F{\mathcal{F}}

\def\X{\mathcal{X}}
\def\Y{\mathcal{Y}}
\def\Z{\mathcal{Z}}
\def\P{\mathcal{P}}

\DeclareMathOperator{\E}{\mathbb{E}}
\DeclareMathOperator{\N}{\mathbb{N}}
\DeclareMathOperator{\R}{\mathbb{R}}

\def\argmin{\mathrm{argmin}}

\def\1{\mathbbm{1}}

\DeclarePairedDelimiter{\p}{\lparen}{\rparen}
\DeclarePairedDelimiter{\abs}{\lvert}{\rvert}
\DeclarePairedDelimiter{\brac}{\lbrack}{\rbrack}

\title{Adversarial VC-dimension and Sample Complexity of Neural Networks}

\author{
  Zetong Qi \\
  Department of Electrical and Computer Engineering\\
  University of Wisconsin - Madison\\
  \texttt{zqi38@wisc.edu} \\
  \and
  \textbf{T.J. Wilder} \\
  Department of Computer Science \\
  University of Wisconsin - Madison\\
  \texttt{twilder@wisc.edu} \\
}

\begin{document}

\maketitle

\begin{abstract}
Adversarial attacks during the testing phase of neural networks pose a challenge for the deployment of neural networks in security critical settings. These attacks can be performed by adding noise that is imperceptible to humans on top of the original data. By doing so, an attacker can create an adversarial sample, which will cause neural networks to misclassify. In this paper, we seek to understand the theoretical limits of what can be learned by neural networks in the presence of an adversary. We first defined the hypothesis space of a neural network, and showed the relationship between the growth number of the entire neural network and the growth number of each neuron. Combine that with the adversarial Vapnik-Chervonenkis(VC)-dimension of halfspace classifiers, we concluded the adversarial VC-dimension of the neural networks with sign activation functions.
\end{abstract}

\section{Introduction}
Machine learning has become the fastest growing area of computer science,
and neural networks are among the most studied among all ML algorithms because of
their impressive performance in areas like image recognition, natural language processing, etc. However, practical neural networks are often vulnerable to adversarial attacks: given an input $x$ and any target label $t$, it is possible to find an $x'$ that is very similar to $x$ but which neural networks will misclassify as the target label $t$. This makes it difficult to apply neural networks in security critical areas like self-driving
cars, malware detection systems, facial recognition systems to unlock devices, etc. Carlini et al. created the state of the art attacks on image recognition systems \cite{DBLP:journals/corr/CarliniW16a} and speech recognition systems \cite{DBLP:journals/corr/abs-1904-05734}. Some have proposed methods, like adversarial training \cite{goodfellow2014explaining}, for creating neural networks that are robust against adversarial attacks.

In this paper, we aim to understand more about neural networks in the presence of a test time adversary. In particular, we try to see if the sample complexity of neural networks is different in the presence of adversaries. We show that, for some simple networks, the sample complexity bound does not change with the introduction of an adversary. We prove this for neural networks with sign activation functions by combining known bounds on sample complexity with recent bounds on the adversarial VC-dimension of halfspace classifiers from Cullina et al. \cite{cullina2018paclearning}

We first describe the adversarial PAC-learning framework. After that, we formally specify our neural network. Finally we prove that the VC-dimension bound is unchanged by the introduction of an adversary.


\subsection{PAC-learning in the presence of an adversary}
In this section, we setup the framework for PAC-learning in the presence of an evasion adversary, which is one that generates and presents the learner with adversarial examples during the test phase but does not interfere with the training.

In our setup, there is an unknown distribution $\P_{\X \times \Y}$. The learner receives training data $S = ((x_1, y_1), \dots , (x_n, y_n)) \sim  \P^n_{\X \times \Y} $ and outputs $\hat{h} \in \H$. The adversary receives data $(x, y) \in \P_{\X \times \Y}$ and outputs some $x' \in N(x)$, where $N(x)$ defines a neighborhood around $x$. The neighborhood of $x$ is defined as $N(x) = \{ x' \in \X : R(x, x') \leq \epsilon \}$ where $R(x, x')$ defines a nearness relationship.

Under this new framework, the adversarial expected risk, $L_P$, is defined as the learner's risk under the true distribution $\P_{\X \times \Y}$ in the presence of an adversary constrained by the nearness relation $R$, and $L_S$ is defined as the adversarial empirical risk with the same adversary
$$L_P(h, R) = \E_{(x,y) \sim \P_{\X \times \Y}} \brac* {\max_{x' \in N(x)} \ell(h(x'), y)}$$
$$L_S(h, R) = \frac{1}{n} \sum_{i=1}^{n} \brac* {\max_{x' \in N(x_i)} \ell(h(x'), y_i)}$$

The goal of the learner is to minimize the adversarial expected risk. Because calculating the expected risk over the distribution $\P_{\X \times \Y}$ is infeasible, the learner instead aims to minimize the adversarial empirical risk. The algorithm that selects the hypothesis $\hat{h}$ that minimizes the adversarial empirical risk from the hypothesis space $\H$, with nearness relation $R$, is called Adversarial Empirical Risk Minimization and we define the objective as follows
$$AERM_{\H, R}(S) = \argmin_{h \in \H}{L_S(h, R)}$$

\subsection{Corrupted hypotheses}
In this section, we describe the concept of a corrupted hypothesis class. The presence of an adversary forces the learner to learn with corrupted hypothesis. Instead of just predicting $\pm 1$, corrupted hypothesis also outputs $\bot$ that means "always wrong".

For $\Y = \{-1, 1\}$, let the corrupted output space be $\widetilde{\Y} = \{-1, 1, \bot \}$. Now we can define the corruption function as a mapping from a hypothesis to the corrupted version $\kappa_R: (\X \mapsto \Y) \mapsto (\X \mapsto \widetilde{\Y})$:
\begin{equation*}
\kappa_R(h) = x \rightarrow \begin{cases}
-1 &\forall x' \in N(x) : h(x') = -1\\
1 &\forall x' \in N(x) : h(x') = 1\\
\bot &\exists x'_0, x'_1 \in N(x) : h(x'_0) \neq h(x'_1)
\end{cases}
\end{equation*}
Using this, we can define the set of corrupted hypotheses as $\widetilde{H} = \{\kappa_R(h): h \in \H\}$. In other words, the hypothesis will predict "always wrong" when the test data $x$ is in the neighborhood where different labels exist.

We also define the loss function $\lambda$, and the loss classes $\F$ and $\widetilde \F$ which are derived from $\H$ and $\widetilde H$ respectively
$$\lambda(h) = \X \times \widetilde{\Y} \rightarrow \{0, 1\}$$
$$\F = \{\lambda(h) : h \in \H\}$$
$$\widetilde{F} = \{\lambda(\widetilde{h}) : \widetilde{h} \in \widetilde{\H}\}$$
We can define the equivalent shattering coefficient in terms of the loss class as
$$\sigma'(\F, i) = \max_{(x',y) \in \X^i \times \Y^i} |\{\p*{f(x'_1, y_1), \dots, f(x'_{i}, y_{i})}: f \in \F\}|$$
Finally, the Adversarial VC-dimension is defined as
$$AVC(\H, R) = \sup\{n \in \N : \sigma'(\lambda(\widetilde{\H}), n) = 2^n\}$$

\subsection{Adversarial VC-dimension of halfspace classifiers}
For the sake of simplicity, we assume that the adversary has standard $\ell_p$ norm-based constraints that are usually imposed on evasion adversaries as described in the literature \cite{DBLP:journals/corr/CarliniW16a, goodfellow2014explaining}. As it turns out, the adversarial VC-dimension for halfspace classifiers corrupted by $\ell_p$ norm-constrained adversary is equal to the standard VC-dimension \cite{cullina2018paclearning}. That is: Let $\H$ be the family of halfspace classifiers of $\X \in \R^d$. Then the adversarial VC-dimension of $\H$ in the presence of an adversary with $\ell_p$ norm-based constraints is $AVC(\H, R) = d+1$.

\section{Adversarial VC-dimension of Neural Networks with Sign Activation Function}
\subsection{Define the Neural Network}
We define a general neural network described by a directed acyclic graph $G = (V, E)$, where all neurons have the same activation function $\sigma(a)$. In a neural network of depth $T$, let $V_0, \dots , V_T$ be the layers of the neural network and let $E_{(t-1, t)}$ be the weights connecting the layers $V_{t-1}$ and $V_t$. Any layer $V_t$ has $\abs* {V_t}$ number of neurons. We can express our neural network's overall hypothesis space as a composition of the hypothesis spaces of each layer, $\H = \H^{(T)} \circ \H^{(T-1)} \circ \dots \circ \H^{(1)}$ where $\H^{(t)} = \{f: \R^{|V_{t-1}|} \mapsto {\R}^{|V_t|}\}$. We analyze the adversarial VC-dimension and sample complexity for this family of hypotheses in the event of an evasion attack.

\subsection{Neural Networks with Sign Activation Function}
To simplify our hypothesis space, we choose the activation function to be $\sigma(a) = \1_{[a>0]}$, and let $\H^{(t)} = \{f: \R^{|V_{t-1}|} \mapsto {\{\pm1\}}^{|V_t|}\}$. With this activation function, each neuron in each layer turns into a halfspace classifier:
$$\forall t \in [T], V_t = sign(\langle E_{(t-1, t)}, V_{t-1} \rangle)$$
Cullina et al. showed that the adversarial VC dimension of a halfspace classifier for $\X = \R^d$ is $d+1$ \cite{cullina2018paclearning}. In the previous section, we showed that the hypothesis class of a neural network $\H$ can be written as a composition of its layers, $\H = \H^{(T)}\circ \dots \circ\H^{(1)}$. The following lemma shows that the growth function of a composition of hypothesis classes is bounded by the products of the growth functions of the individual classes.
\begin{lemma}
Let $\F_1$ be a set of functions from $\X$ to $\Z$ and let $\F_2$ be a set of functions from $\Z$ to $\Y$. Let $\H = \F_2 \circ \F_1$ be the composition class. That is, $\forall f_1 \in \F_1$ and $f_2 \in \F_2 : \exists h \in \H \text{ s.t. } h(x) = f_2(f_1(x))$. The growth function of $\H$, $\tau_{\H}(m)$, is bounded by $\tau_{\H}(m) \leq \tau_{\F_1}(m)\tau_{\F_2}(m)$ \\

\textit{Proof:} \\
\normalfont{Let $C = \{c_1, \dots, c_m\} \subseteq \X$. Then}
\begin{align*}
    |\H_C| &= |\{f_2(f_1(c_1)), \dots, f_2(f_1(c_m)) : f_1 \in \F_1, f_2 \in \F_2\}| \\
    &= \abs*{\bigcup_{f_1 \in \F_1} \{(f_2(f_1(c_1)), \dots, f_2(f_1(c_m))) : f_2 \in \F_2\}} \\
    &\leq \abs{\F_{1C}} \cdot \tau_{\F_2}(m) \\
    &\leq \tau_{\F_1}(m) \tau_{\F_2}(m) \\
    \tau_{\H}(m) &\leq \tau_{\F_1}(m) \tau_{\F_2}(m) \hspace{5cm} \square
\end{align*}
\end{lemma}
Therefore, the growth function of the hypothesis space of neural networks is bounded by:
$$\tau_{\H}(m) \leq \prod_{t=1}^{T}\tau_{\H^{(t)}(m)}$$
In addition, each $\H^{(t)}$ can written as a product of individual neurons, $\H^{(t)} = \H^{(t, 1)} \times \dots \times \H^{(t, \abs* {V_t})}$, where each $\H^{(t, j)}$ is a halfspace classifier: $\H^{(t, j)} = \{f: \R^{|V_{t-1}|} \mapsto {\{\pm1\}}\}$. The following lemma shows that the growth function of the Cartesian product class is bounded by the products of the growth functions of the individual classes:
\begin{lemma}
For $i = 1, 2$, let $\F_i$ be a set of functions from $\X$ to $\Y_i$. Define $\H = \F_1 \times \F_2$ to be the Cartesian product class. That is, $\forall f_1 \in \F_1$ and $f_2 \in \F_2: \exists h \in \H \text{ s.t. } h(x) = (f_1(x), f_2(x))$. The growth function of $\H$ is bounded by: $\tau_{\H}(m) \leq \tau_{\F_1}(m)\tau_{\F_2}(m)$

\textit{Proof:} \\
\normalfont{Let $C = \{c_1, \dots, c_m\} \subseteq \X$. Then}
\begin{align*}
    |\H_C| &= |\{((f_1(c_1), f_2(c_1)), \dots, (f_1(c_m), f_2(c_m))) : f_1 \in \F_1, f_2 \in \F_2\}| \\
    &= |\{((f_1(c_1), \dots, f_1(c_m)), (f_2(c_1), \dots, f_2(c_m))) : f_1 \in \F_1, f_2 \in \F_2\}| \\
    &= |\F_{1C} \times \F_{2C}| \\
    &= |\F_{1C}| \cdot |\F_{2C}| \\
    \tau_{\H}(m) &\leq \tau_{\F_1}(m) \tau_{\F_2}(m) \hspace{5cm} \square
\end{align*}
\end{lemma}

Therefore, the growth function of each layer can be bounded by:
$$\tau_{\H^{(t)}}(m) \leq \prod_{i=1}^{\abs* {V_t}} \tau_{\H^{(t, i)}}(m)$$
Combining these two lemmas, the growth function of the entire neural network is bounded by:\\
$$\tau_{\H}(m) \leq \prod_{t=1}^T \prod_{i=1}^{\abs* {V_t}} \tau_{\H^{(t, i)}}(m)$$
Since the adversaries are only able to change the inputs, we only corrupt the first layer by introducing an adversary. However, since the adversarial VC-dimension of halfspace classifiers of dimension $d$ is $d+1$, the same as the regular VC-dimension, we can say that the $i$th neuron in \textit{all} hidden layers have an effective VC-dimension of $d_{t,i}$, where $d_{t,i}$ is the number of edges that are going into the $i$th neuron of the $t$th layer, assuming that one edge accounts for the bias term.

Using this alongside our growth function bound, we can use Sauer's lemma to show that:\\
$$\tau_{\H}(m) \leq \prod_{t=1}^T \prod_{i=1}^{\abs*{V_t}} \p*{\frac{em}{d_{t,i}}}^{d_{t,i}} \leq \prod_{t=1}^T \prod_{i=1}^{\abs*{V_t}} \p*{em}^{d_{t,i}} = \p*{em}^{\p*{\sum_{t=1}^{T} \sum_{i=1}^{\abs*{V_t}}d_{t,i}}}$$
Notice that ${\sum_{t=1}^{T} \sum_{i=1}^{\abs* {V_t}}d_{t,i}}$ is just the number of edges in the neural network, therefore we have:\\
$$\tau_{\H}(m) \leq \p*{em}^{\abs* E}$$
Let there be a set of size $m$ that is shattered by the neural network. Therefore the growth number $\tau_{\H}(m) = 2^m$. Combining this with $\tau_{\H}(m) \leq \p*{em}^{\abs* E}$, we have that:
$$2^m \leq \p*{em}^{\abs* E}$$
The following lemma shows that $m$ must be $\O(\abs* E \log(\abs* E))$ in order to satisfy the inequality.

\begin{lemma}
If a neural network's hypothesis space $\H$ has $\abs* E$ number of parameters, let $m$ be the size of the set $\H$ shatters. If the inequality $2^m \leq \p*{em}^{\abs* E}$ is satisfied, then $\H$ has sample complexity of $\O(\abs* E \log(\abs* E))$.\\

\textit{Proof:} \\
\begin{align*}
    2^m &\leq (em)^{|E|}\\
    m \log(2) &\leq |E| \log(em) \\
    m &\leq \frac{|E|}{\log(2)} \log(em) \\
    em &\leq \frac{e|E|}{\log(2)} \log(em) \\
\noalign{\normalfont{Lemma A.1 in \cite{Shalev-Shwartz:2014:UML:2621980} states:
"for $a \geq 0$, if $x \geq 2a\log(a)$ then $x \geq a\log(x)$". We can take the contrapositive to get: "if $x < a\log(x)$ then $x < 2a\log(a)$" and apply it}} \\
    em &< \frac{2e|E|}{\log(2)} \log\p*{\frac{e|E|}{\log(2)}} \\
    m &< \frac{2|E|}{\log(2)} \log\p*{\frac{e|E|}{\log(2)}} \\
\noalign{\normalfont{Ignoring the constants, we have:s}} \\
    m &\leq \O(|E|\log |E|) \hspace*{2cm} \square
\end{align*}
\end{lemma}

Putting all of these pieces together, we now have a bound on the adversarial VC-dimension for our network.

\begin{theorem}
For a neural network with sign activation functions, the adversarial VC-dimension has the same bound as the regular VC-dimension.

\textit{Proof:}\\
\normalfont{From Lemma 3, we can see that the AVC-dimension of neural networks with sign activation function is $\O(|E| \log |E|)$.
We know from Theorem 20.6 in \cite{Shalev-Shwartz:2014:UML:2621980} that the regular VC-dimension of neural networks with sign activation function is also $\O(|E| \log |E|)$.}
\end{theorem}

\section{Extending corrupted hypotheses}
One big issue with the corrupted hypothesis class as defined by Cullina et al. is that it limits the expressiveness of the hypotheses by requiring $\pm 1$ output \cite{cullina2018paclearning}. In order to overcome this issue, we introduce two generalizations of the corrupted hypothesis class. The first is a generalization to multi-class corrupted hypotheses.
\begin{align*}
    \kappa_R(h) = x \rightarrow \begin{cases}
        h(x) &\forall x' \in N(x) : h(x) = h(x')\\
        \bot &\exists x'_0, x'_1 \in N(x) : h(x'_0) \neq h(x'_1)
    \end{cases}
\end{align*}

This is more of a generalization in notation than anything else because we simply output the normal label of $x$ for the hypothesis if it agrees on all the neighbors of $x$, and reject it if it doesn't. In the $\pm 1$ case, this version simplifies down to the original form of the equation.

However, this version is still impractical for problems without a reasonable number of outputs or for any problem which has more complex relationships between the outputs. To solve these problems, we introduce our second new form of the corrupted hypothesis class which we call the continuous corruption class.
\begin{align*}
    \kappa_{R,d}(h) = x \rightarrow \begin{cases}
        h(x) &\forall x' \in N(x) : d(h(x), h(x')) \leq \delta\\
        \bot &\exists x'_0, x'_1 \in N(x) : d(h(x'_0), h(x'_1)) > \delta
    \end{cases}
\end{align*}

Here, $d$ defines a distance metric on $\Y$ which is analogous to $R$, the distance metric on $\X$. Instead of looking for perfect agreement among all the neighbors, we instead allow normal predictions on $x$, when any of the neighbors of $x$ would have sufficiently similar outputs. This formulation now allows us to describe the corrupted version of any hypothesis class, instead of only binary ones. It should be possible to theoretically describe how easy it is to corrupt any hypothesis class by comparing the normal and corrupted versions.

This formulation directly relates to our own problem of discovering VC-dimension of neural networks in adversarial environments. The original framework restricted us to $\pm 1$ output for the network and, for our proof of the bound, even for each node of the network. By using the continuous corruption class, we can instead look at adversarial elements of continuous functions, including many other common activation functions such as Sigmoids, Tanh, or ReLU. All of these are actually used in practice, while our version using sign activation is primarily a theoretical model.

Beyond accommodating for continuous functions, this generalization also allows for more interesting relationships between the output. For example, when classifying images for self-driving cars, it is a big problem if an adversary can make your model see a car instead of a human, but probably less of a problem if your model is made to see a car instead of a truck. Obviously there are may be problems either way, but when you have a lot of classes or continuous outputs, then intelligently picking your distance metric could help make the model robust without relying on perfect predictions.

\section{Conclusion}
We analyzed the adversarial VC-dimension and sample complexity for neural networks with sign activation function and showed that it could achieve the same bound as the non-adversarial case. Though Cullina et. al. did show that the AVC-dimension could be arbitrarily larger or smaller than the ordinary VC-dimension, we have shown that there exist learners, including at least some neural networks, which may be highly resilient to adversarial attacks given sufficient training data. In practice however, we find that neural networks are often highly susceptible to adversarial attacks. This may be because of the number of training samples used in practice is usually significantly less than the sample complexity, and the learned model is overfitting to the limited training set so it doesn't generalize well to the true distribution.

A natural extension to our work, which may help to show if that is true, is to find the adversarial VC-dimension of neural networks with some other activation functions such as Sigmoid, Tanh or ReLU. These are of particular interest because they are some of the most widely used in practice. The difficulty with the analysis of those activation functions arises from the fact that they are continuous and map $\R \mapsto \R$. Because each neuron has real output, it becomes hard to use the same proof techniques, because we cannot look at the AVC-dimension for a single neuron. That being said, there are many other methods which could potentially be used to prove these bounds. These include existing methods which bound the sample complexity for neural networks with sigmoid activation functions. For some of these functions, we may also be able to look at an adversarial Rademacher complexity and bound the sample complexity that way. We leave the exploration of these methods to future work.

\medskip 

\bibliographystyle{unsrt}
\bibliography{references.bib}

\begin{thebibliography}{1}

\bibitem{DBLP:journals/corr/CarliniW16a}
Nicholas Carlini and David~A. Wagner.
\newblock Towards evaluating the robustness of neural networks.
\newblock {\em CoRR}, abs/1608.04644, 2016.

\bibitem{DBLP:journals/corr/abs-1904-05734}
Hadi Abdullah, Washington Garcia, Christian Peeters, Patrick Traynor, Kevin
  R.~B. Butler, and Joseph Wilson.
\newblock Practical hidden voice attacks against speech and speaker recognition
  systems.
\newblock {\em CoRR}, abs/1904.05734, 2019.

\bibitem{goodfellow2014explaining}
Ian~J. Goodfellow, Jonathon Shlens, and Christian Szegedy.
\newblock Explaining and harnessing adversarial examples, 2014.

\bibitem{cullina2018paclearning}
Daniel Cullina, Arjun~Nitin Bhagoji, and Prateek Mittal.
\newblock Pac-learning in the presence of evasion adversaries, 2018.

\bibitem{Shalev-Shwartz:2014:UML:2621980}
Shai Shalev-Shwartz and Shai Ben-David.
\newblock {\em Understanding Machine Learning: From Theory to Algorithms}.
\newblock Cambridge University Press, New York, NY, USA, 2014.

\end{thebibliography}

\end{document}